\documentclass[12pt, fleqn]{article}

\usepackage[english]{babel}

\usepackage{xcolor}
\usepackage{hyperref}
\definecolor{linkcolor}{HTML}{000000}
\definecolor{citecolor}{HTML}{0B6B1A}
\definecolor{urlcolor}{HTML}{0B2A5B}
\hypersetup{pdfstartview=FitH, citecolor=citecolor, linkcolor=linkcolor, urlcolor=urlcolor, colorlinks=true}

\usepackage[pdftex]{graphicx}
\graphicspath{{.\\}}
\DeclareGraphicsExtensions{.pdf,.png,.jpg}

\usepackage{ragged2e}
\usepackage{microtype}

\begin{document}

\title{Natural Way to Overcome Catastrophic Forgetting in Neural Networks}
\author{\bf Alexey Kutalev}

\date{\small JSC InfoWatch, Moscow, Russia}
\maketitle

Code: \texttt{\href{https://github.com/aakutalev/wva}{https://github.com/aakutalev/wva}}

\abstract{Not so long ago, a method was discovered that successfully overcomes the catastrophic forgetting of neural networks. Although we know about the cases of using this method to preserve skills when adapting pre-trained networks to particular tasks, it has not yet obtained widespread distribution. In this paper, we would like to propose an alternative method of overcoming catastrophic forgetting based on the total absolute signal passing through each connection in the network. This method has a simple implementation and seems to us essentially close to the processes occurring in the brain of animals to preserve previously learned skills during subsequent learning. We hope that the ease of implementation of this method will serve its wide application.}

\section{Introduction}

The problem of {\it catastrophic forgetting} \cite{c3,c4} affects many tasks in modern machine learning, making it difficult to preserve the skills of pre-trained neural networks (NN) during further training. For example, when adapting products that are state-of-the-art in the ML industry (such as BERT, GPT, etc.) to the application tasks of a particular developer. In this case, the neural network is retrained on specialized datasets without access to the original datasets, on which the primary training of NN was performed.

The matter of the problem of catastrophic forgetting is that a neural network trained on some dataset {\bf A} quickly loses the skill acquired by learning on {\bf A} while further training on another training set {\bf B} in the absence of set {\bf A}. Such a behavior does not at all resemble the behavior of animals, which are capable of retaining learned skills for a long time when learning other tasks. This discrepancy motivates us to look for ways to overcome this problem.

From neurophysiological and machine learning \cite{c1,c2,c3,c4,c6} researches we can conclude that the root of the problem is that learning on the set {\bf B} changes the connections in neural network that are important for preserving the skill acquired during training on set {\bf A}. It would be logical to try to preserve in some way the connections important for the learned skill  during further training on another training data sets. The method which is following this logic have been proposed recently to overcome the problem of catastrophic forgetting in NN \cite{c6}. However, so far we have not observed widespread use of this method.

In this article, we would like to propose a method for overcoming the problem of catastrophic forgetting, which is based on the method of elastic weights consolidation (EWC) proposed by Kirkpatrick et al. \cite{c6}. Unlike the EWC, the ``significance'' of the connection weight in our algorithm is based not on the diagonal elements of the Fisher information matrix, but on the total absolute signal that passed through the connection during processing by the network of training examples after the completion of the training cycle. Like EWC, our approach has a linear computational cost in terms of the number of network parameters and the number of examples in the training set. However, it avoids the building of an additional computational graph (for calculating the diagonal of the Fisher matrix) and can be integrated directly into the learning process of a neural network.

We will also consider the cases for which the method of elastic weight consolidation fails.

\section{Results}

The algorithm proposed by Kirkpatrick et al. \cite{c6} shows the impressive ability to preserve the skill gained on one training dataset while training on other datasets. At the same time, the mechanism for solving the problem of catastrophic forgetting used by the animal brains probably has a different nature: it is difficult to imagine that each individual dendrite of the animal's nervous system is able to calculate the corresponding element of the Fisher matrix. This induces us to look for other methods of preserving the skill in the neural network.

Following the logic of the elastic weight consolidation method (EWC), each connection weight $w_i$ in the NN is matched by its ``significance'' for the skill acquired during training on the first training set {\bf A}. The ``significance'' of the weight $w_i$ shows how much the change of the weight $w_i$ will be penalized when training on the next training set {\bf B}. That is, a change in the weight of $w_i$ is penalized the more, the more its ``significance'' is.

The penalty is achieved by adding the regularizer to the loss function:
$$
L = L_B + \frac{\lambda}{2} \sum_i F_i (w_i - w^*_{A,i})^2, 
$$
where $L_B$ is the loss for training on dataset {\bf B} only, $w^*_{A, i}$ is the neural network weights after training on the dataset {\bf A}, and as the ``significance'' of the $i$-th weight here acts $F_i$ -- corresponding diagonal element of the Fisher information matrix. Theoretical basis and the details about $F_i$ calculations can be found at original article \cite{c6}.

As an alternative to $F_i$, we suggest to use a different value characterizing the ``significance'' of connection in a neural network. Namely, the total absolute signal that passed through the connection during processing by the trained network of all the examples from the training dataset {\bf A}:
$$
S^w_{A,i} = \frac{1}{n} \sum_k |x_{k,i} w_{A,i}|,
$$
where $x_{k, i}$ is the signal applied to the input of the $i$-th connection when the neural network processes the $k$-th sample of dataset, $n$ is the number of samples in the dataset {\bf A}, and $w_{A, i}$ is the weight of the $i$-th connection in a neural network trained on {\bf A}. As the "significance" of the $j$-th bias of the neural network, we take the value:
$$
S^b_{A,j} = \frac{1}{n} \sum_k |y_{k,j}|,
$$
where $y_{k, j}$ is the output signal (activation) of the $j$-th neuron, to the adder of which the bias $b_j$ belongs to.

Thus, the loss function while training on the subsequent dataset {\bf B} becomes:
$$
L = L_B + \frac{\lambda}{2} \sum_i S^w_{A,i} (w_i - w^*_{A,i})^2 +
\frac{\lambda}{2} \sum_j S^b_{A,j} (b_j - b^*_{A,j})^2,
$$ 
where $S^w_{A, i}$ and $S^b_{A, j}$ ``significance'' of corresponding weights and biases obtained after training network on the dataset {\bf A}.

For further training on the dataset {\bf C}, we must add the corresponding regularization members. Then the loss function for training on {\bf C} takes form:
$$
L = L_C + \frac{\lambda}{2} \sum_i (S^w_{A,i}+S^w_{B,i}) (w_i - w^*_{B,i})^2 +
\frac{\lambda}{2} \sum_j (S^b_{A,j}+S^b_{B,j}) (b_j - b^*_{B,j})^2.
$$
That is, with sequential training, the ``significance'' of weights and biases accumulates:
$$
S_{AB,i} = S_{A,i} + S_{B,i},
$$
and so on for all subsequent training datasets. Correctness of ``significance'' accumulation against adding separate regularization term for each next training (as proposed by Kirkpatrick et al. \cite{c6}) is provided in \cite{c9}.

When learning using the gradient descent optimization, the change in the network weights at each training step looks like:
$$
w'_i = w_i - \alpha \nabla_i,
$$
where $\alpha$ is the learning rate, $\nabla_i = \frac{\partial L}{\partial w_i}$ is the gradient of the loss function by the connections weights. Instead of adding a regularizer to the loss function $L$, we can weaken the gradient of the loss function at each training step in proportion to the value of the ``significance'' of the connection $S^w_i$ accumulated during training on the previous datasets. In the case of zero ``significance'' of weights, that is, when training on the first data set, the gradient should have a coefficient of $1$ and must decrease as the ``significance'' of the connection grows. To fulfill these conditions, we use the factor $\frac{1}{1+\lambda S^w_i}$:
$$
w'_i = w_i -  \frac{\alpha}{1+\lambda S^w_i}\nabla_i.
$$
For training biases, similarly:
$$
b'_j = b_j -  \frac{\alpha}{1+\lambda S^b_j}\nabla_j.
$$
We named this approach the method of Weight Velocity Attenuation (WVA). In the strict mathematical sense, WVA is the limiting case of applying the EWC to each individual batch in the training process as to a separate dataset in sequential training.

In the WVA method the diagonal elements of the Fisher information matrix  $F$ can also act as ``significance'' of corresponding weights and biases instead of the total absolute signal $S$. Thus, we get four possible combinations of methods for consolidating weights and attenuating the gradient with the used ``significance'' of weights based on the Fisher matrix and the total absolute signal.

\begin{figure}[!h]
\centering
\parbox[t]{.7\textwidth}{
%\centering
\includegraphics[width=.7\textwidth]{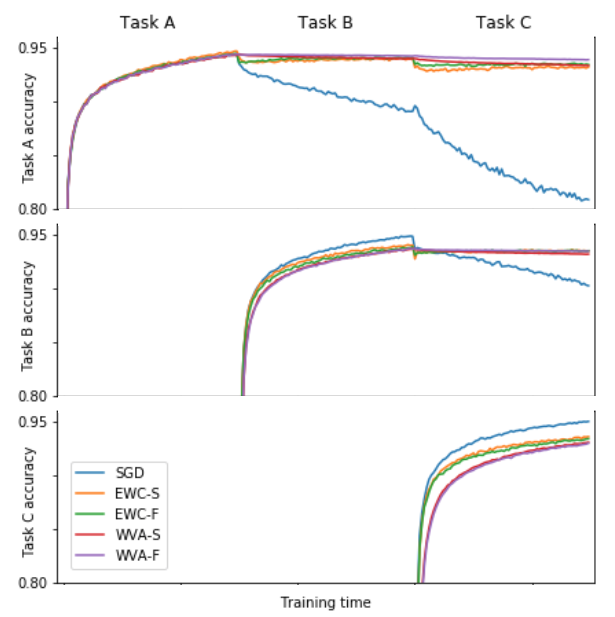}
\caption{\footnotesize \textsf{Accuracy on the corresponding training set (vertical) during sequential training on three training sets (horizontal) using methods: SGD -- simple gradient descent, EWC-S -- elastic weight consolidation based on the total absolute signal, EWC-F -- elastic weight consolidation based on the diagonal elements of the Fisher matrix, WVA-S -- weight velocity attenuation based on the total absolute signal, WVA-F -- weight velocity attenuation based on the diagonal elements of the Fisher matrix. The graph was averaged over 10 passes.}}
\label{wvafig1}}
\end{figure}
To test these methods, we conducted experiments on the sequential training of deep neural networks with a different number of fully connected layers on several training datasets. Each of these sets was obtained from the MNIST data set by randomly permutating inputs in the same way for all examples in the set, similar to how it was done by Kirkpatrick et al. \cite{c6}.

\begin{figure}[!h]
\centering
\parbox[t]{.7\textwidth}{
%\centering
\includegraphics[width=.7\textwidth]{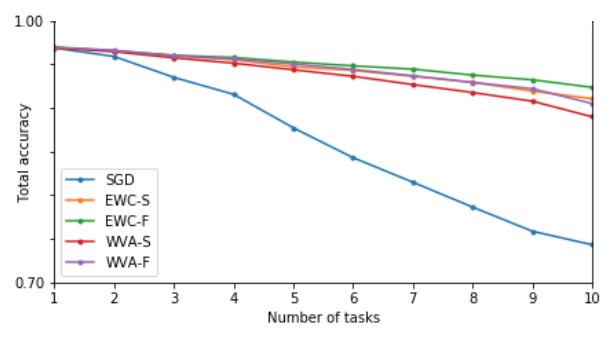}
\caption{\footnotesize \textsf{Degradation of total accuracy on all trained sets in sequential training. The graph was averaged over 10 passes.}}
\label{wvafig2}}
\end{figure}

Our experiments showed that all four of the above methods demonstrate almost the same ability of NN to retain skills during sequential training of 
several training sets -- see Figures \ref{wvafig1} and \ref{wvafig2}.
According to our observations, ceteris paribus, methods of weight velocity attenuation are only slightly inferior to the methods of elastic weight consolidation.
This is intuitively predictable, since elastic weight consolidation does not allow ``significant'' network weights to go far from the values of fixing, while weight velocity attenuation allows more ``significant'' weights to move away from the values of fixing arbitrarily far. It just happens much slower than for less ``significant'' weights. Also, if you choose a method with the parameter $\lambda_1$ and elements of the Fisher matrix as the ``significance'' of the weights, then you may find such a coefficient $\lambda_2$ that the same method based on the total absolute signal with the parameter $\lambda_2$ shows the same ability of skill preservation during sequential training on several training sets.

\section{Observations}

\begin{figure}[h]
\centering
\parbox[t]{.7\textwidth}{
%\centering
\includegraphics[width=.7\textwidth]{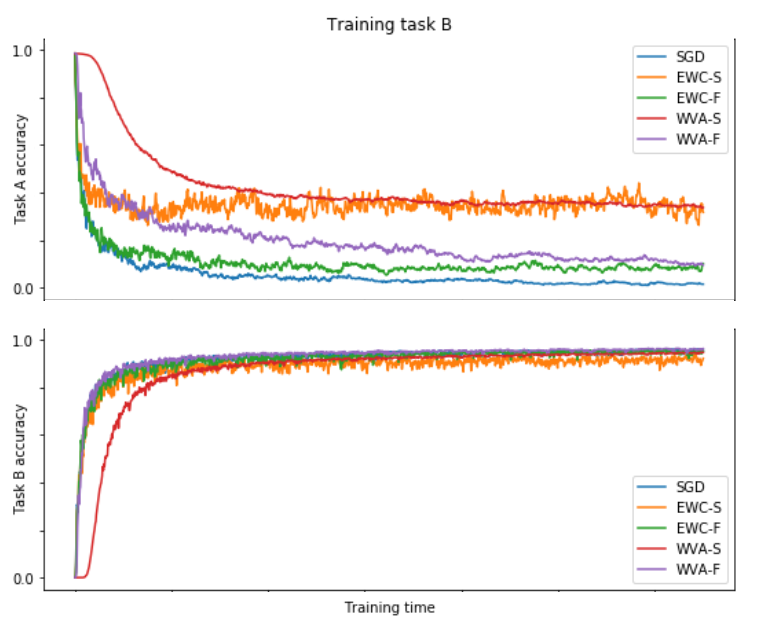}
\caption{\footnotesize \textsf{Accuracies on the datasets {\bf A} and {\bf B} after training on dataset {\bf A} while training on dataset {\bf B} using various skills preservation methods. The graph is averaged over 10 passes.}}
\label{wvafig3}}
\end{figure}

All the methods described above solve the problem of catastrophic forgetting of neural networks only in the case when each of the datasets during sequential training includes examples with activation of each of the outputs of neural network. If the set contains examples of only part of the classes recognized by the neural network, then catastrophic forgetting quickly destroys previous skills with sequential training, even using methods of elastic weight consolidation or weight velocity attenuation. An illustration of the problem can be seen in Figure \ref{wvafig3}, which shows the degradation of accuracy during training on the second dataset {\bf B} using various methods of skill preservation following training on the first set {\bf A}.

The training sets {\bf A} and {\bf B} are obtained from MNIST: {\bf A} contains only examples with digits $0$-$4$, {\bf B} contains examples with digits $5$-$9$. You can see that only when using the EWC-S and WVA-S methods, based on the total absolute signal, it is possible to save about 35\% of the skill obtained during training on {\bf A}. And when using the method of weight velocity attenuating by the total absolute signal (WVA-S), it is possible to get the total accuracy of about 75\% using an early stopping (however, it requires the 
usage of test part of dataset {\bf A} when learning on {\bf B}). 
In any case, the accuracy on the set {\bf A} degrades significantly more than when using ``complete'' training sets, which include examples of each of the classes recognized by the neural network.

Such behavior when training NN is very different from the behavior of animal brain learning. This can be shown with a simple example: a cat can learn to walk and move its ears sequentially and independently. While learning to walk, the outputs of its brain that control the ears are not activated. And while learning to move the ears, the outputs that control the paws are not activated. However, with learning to move the ears, the walking skill does not degrade significantly.

\section{Discussion}

Since the proposed methods, based on the total absolute signal passed through the connection, help to maintain skills during sequential training, allows us to hypothesize that learning in the brains of humans and animals occurs according to a similar pattern. First, the network learns from the examples provided. Then, when a satisfactory result is obtained on several examples provided to the network, the most important connections between neurons are consolidated. This consolidation can occur, for example, by releasing a supporting hormone, which makes less ductile the connections with a greater number of spikes passed through. As per research done by Hiroaki Wake et al. \cite{c7} and Daniel J. Miller et al. \cite{c8}, similar behavior is demonstrated by hormone myelin. That is, myelination of connections that have let through the largest number of spikes (signal transmissions) occurs.

Based on the conducted experiments with slowing down the training of connections, we can make the assumption that learning the animal's brain in a specific task (training set) leads to the creation of an entire consolidated subnetwork in its brain. Thus, as a result of sequential training in several tasks, the brain becomes a complex of subnetworks, each of which is trained to solve a separate task.


\newpage

\begin{thebibliography}{99}

\bibitem{c1} McCloskey M, Cohen NJ (1989) Catastrophic interference in connectionist networks: The sequential learning problem. \textit{The Psychology of Learning and Motivation}, ed GH Bower (Academic, New York), Vol 24, pp 109-165

\bibitem{c2} McClelland JL, McNaughton BL, O'Reilly RC (1995) Why there are complementary learning systems in the hippocampus and neocortex: Insights from the successes and failures of connectionist models of learning and memory. \textit{Psychol Rev} 102(3): 419-457.

\bibitem{c3} French RM (1999) Catastrophic forgetting in connectionist networks. \textit{Trends Cognit Sci} 3(4):128-135.

\bibitem{c4} Goodfellow IJ, Mirza M, Xiao D, Courville A, Bengio Y (2015) An empirical investigation of catastrophic forgetting in gradient-based neural networks. arXiv:1312.6211.

\bibitem{c6} James Kirkpatrick and Razvan Pascanu and Neil Rabinowitz and Joel Veness and Guillaume Desjardins and Andrei A. Rusu and Kieran Milan and John Quan and Tiago Ramalho and Agnieszka Grabska-Barwinska and Demis Hassabis and Claudia Clopath and Dharshan Kumaran and Raia Hadsell (2016) Overcoming catastrophic forgetting in neural networks. arXiv:1612.00796.

\bibitem{c7} Hiroaki Wake, Philip R. Lee, R. Douglas Fields (2011) Control of Local Protein Synthesis and Initial Events in Myelination by Action Potentials. \textit{Science}  16 Sep 2011: Vol. 333, Issue 6049, pp. 1647-1651

\bibitem{c8} Daniel J. Miller, Tetyana Duka, Cheryl D. Stimpson, Steven J. Schapiro, Wallace B. Baze, Mark J. McArthur, Archibald J. Fobbs, Andre M. M. Sousa, Nenad Sestan, Derek E. Wildman, Leonard Lipovich, Christopher W. Kuzawa, Patrick R. Hof, and Chet C. Sherwood (2012) Prolonged myelination in human neocortical evolution. \textit{National Academy of Sciences} https://www.pnas.org/content/early/2012/09/18/1117943109 

\bibitem{c9}	Huszar F. (2018) Note on the quadratic penalties in elastic weight consolidation. \textit{Proceeding of the National Academy of Science}, 2018. Vol. 115, issue 11. pp. 2496–2497. DOI: 10.1073/pnas.1717042115

\bibitem{c10} Zenke F., Poole B., and Ganguli S. (2017) Continual learning through synaptic intelligence \textit{Proceedings of the 34th International Conference on Machine Learning (ICML)}

\bibitem{c11} Aljundi R., Babiloni F., Elhoseiny M., Rohrbach M. and Tuytelaars T. (2018) Memory aware synapses: Learning what (not) to forget. \textit{The European Conference on Computer Vision (ECCV)}, September 2018.

\bibitem{c12} Thangarasa V., Miconi T., Taylor G.W. (2020) Enabling Continual Learning with Differentiable Hebbian Plasticity.  arXiv:2006.16558. \textit{International Joint Conference on Neural Networks (IJCNN)}, Glasgow, United Kingdom, 2020, pp. 1-8, DOI: 10.1109/IJCNN48605.2020.9206764.

\end{thebibliography}
\end{document}